\pdfoutput=1

\documentclass[11pt]{article}

\usepackage{acl}

\usepackage{times}
\usepackage{latexsym}

\usepackage[T1]{fontenc}

\usepackage[utf8]{inputenc}

\usepackage{microtype}

\usepackage{CJKutf8}

\usepackage{xspace}

\newcommand{\hide}[1]{} 

\usepackage{hyperref}
\usepackage{array}
\usepackage{multirow}
\usepackage{makecell}
\usepackage{amsmath}
\usepackage{amssymb}
\usepackage{verbatim}
\usepackage{amsfonts}
\usepackage{graphicx}
\usepackage{color}
\usepackage{float}
\usepackage{bm}
\usepackage{arydshln}
\usepackage[noend]{algpseudocode}
\usepackage{tikz}
\usepackage{booktabs}
\usepackage[linesnumbered, ruled, vlined]{algorithm2e}
\usepackage{threeparttable}
\usepackage{lipsum}

\usepackage[figuresleft]{rotating}

\DeclareMathOperator*{\argmax}{arg\,max}
\DeclareMathOperator*{\argmin}{arg\,min}

\title{MetaPrompting: Learning to Learn Better Prompts}

\author{Yutai Hou\textsuperscript{*}, Hongyuan Dong\textsuperscript{*}, Xinghao Wang, Bohan Li, Wanxiang Che\textsuperscript{\dag}\\
Research Center for Social Computing and Information Retrieval \\
Harbin Institute of Technology, China \\
\texttt{\{ythou, hydong, xhwang, bhli, car\}@ir.hit.edu.cn}}

\begin{document}
\maketitle
\footnotetext{* Equal contribution.}
\footnotetext{\dag Email corresponding}
\begin{abstract}
Prompting method is regarded as one of the crucial progress for few-shot nature language processing.
Recent research on prompting moves from discrete tokens based ``hard prompts'' to continuous ``soft prompts'', which employ learnable vectors as pseudo prompt tokens and achieve better performance.
Though showing promising prospects, these soft-prompting methods are observed to 
rely heavily on good initialization to take effect. 
Unfortunately, obtaining a perfect initialization for soft prompts requires understanding of inner language models working and elaborate design, which is no easy task and has to restart from scratch for each new task.
To remedy this, we propose a generalized soft prompting method called MetaPrompting, 
which adopts the well-recognized model-agnostic meta-learning algorithm 
to automatically find better 
prompt initialization that facilitates fast adaptation to new prompting tasks.
Extensive experiments show MetaPrompting
tackles soft prompt initialization problem and brings significant improvement on four different datasets
(over $7$ points improvement in accuracy for 1-shot setting), achieving new state-of-the-art performance. 
\end{abstract}

\section{Introduction}
Enabling models to learn from a few labeled examples, i.e., Few-Shot Learning (FSL), is one of the key steps toward more human-like artificial intelligence. Recently, taking advantage of large-scale Pretrained Language Models (PLM)~\cite{brown2020language}, prompting-based methods achieve impressive results for few-shot learning of Natural Language Processing (NLP)~\cite{gao2020making,liu2021makes,zhao2021calibrate}.

Prompting-based methods insert a piece of text, i.e. prompts, to the input examples, so that the few-shot task can be formulated as a (masked) language modeling problem. 
For example, say we want to classify the sentiment of the book review ``I will never read it again.'', we can append a prompt ``It was'' to the sentence, getting ``I will never read it again. It was''. It is natural to expect a higher probability from the PLM to generate ``terrible'' than ``great'' then. 
Such converting bridges the gap between pre-training and target tasks. Consequently, it has better transferability and less dependence on target task data.

\begin{figure}[t]
\centering
\begin{tikzpicture}
\draw (0,0 ) node[inner sep=0] {\includegraphics[width=0.9 \columnwidth, trim={0cm 0cm 0cm -1cm}, clip]{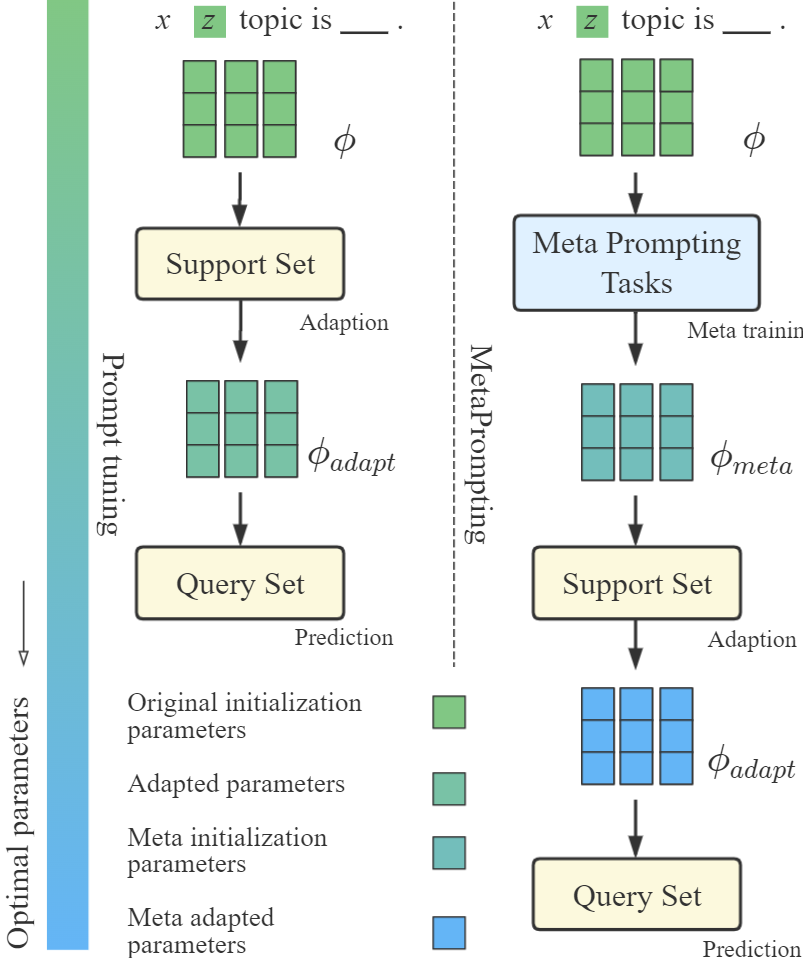}};
\end{tikzpicture}
\caption{Comparison between conventional soft-prompting method (left) and proposed MetaPrompting (right). 
$x$ denotes the query sentence, and $z$ is learnable pseudo tokens in soft prompts. $\phi$ represents all trainable parameters. 
MetaPrompting exploits optimization-based meta-learning to find an initialization $\phi_\text{meta}$ that facilitates better and faster adaptation to new tasks.
\vspace*{-4mm}
}\label{fig:intro}
\end{figure}

The performance of prompting methods is found to be greatly affected by the design of prompts \cite{gao2020making}.
That is, a good prompt makes significant difference.
Early attempts take manually-designed prompts or search prompts automatically. 
\citet{schick2020coling} and \cite{schick2021eacl} explore to automatically identify label words.
In pursuit of better performance compared to hand-picked prompts, \citet{gao2020making} proposes LM-BFF to search both prompt templates and label words. 
AutoPrompt \cite{shin2020autoprompt} leverages gradient-based searching to find out the best prompts.
These prompts consist of discrete tokens, which limits the prompt search space.
To further liberate the potential of prompts, recent works employ learnable vectors as prompt content and learn optimal prompts in continuous space, which is so-called ``soft prompts'' \cite{P-tuning,li2021prefix}.
Since they no longer require prompts to be composed of real words, these methods greatly expand the possibilities of prompts and thus achieve better performance \cite{liu2021pre}. 

However, despite the promising prospects of soft prompts, learning a good prompt is still far from trivial. 
Because soft-prompts search for optimal solutions in an infinite continuous space, the choice of the starting point for the search (i.e., prompt initialization) becomes crucial.
Soft-prompt is observed to be more sensitive to different initialization than discrete prompts in low data setting \cite{li2021prefix,liu2021pre}. 
Unfortunately, creating a perfect prompt initialization requires both understanding of LMs' inner workings and trial-and-error.
\citet{lester2021power} initialize soft prompt with the token embeddings of hand-crafted prompt directly.
\citet{zhong2021fact} search discrete tokens as better initialization, which shows better performance.
What's worse is that these initializations are task-bounded.
Every time we confront a new task, the costly process of initialization design has to start from scratch.

In this paper, to tackle the above issues, we let loose the prompt design of a specific task, but instead focus on obtaining a task general prompt initialization that facilitates faster and better adaptation to new prompting tasks.
Recently proposed optimization-based meta-learning algorithms, such as MAML~\cite{MAML} and  Reptile~\cite{Reptile}, achieve better adaption by learning a parameter initialization. Following their essence, we tackle soft prompt initialization problem by proposing MetaPrompting, which is a generalized soft prompting method powered by meta-learning algorithms. MetaPrompting learns general meta-knowledge from source domain tasks to form a better soft prompt initialization, and thus adapts faster and better across various target domain tasks (See Figure \ref{fig:intro}). Extensive experiments show that MetaPrompting achieves promising performance with desired robustness. 


We summarize the main contribution of this paper as follows:

(1) We propose a novel prompting method MetaPrompting, which employs optimization-based meta-learning algorithm to find adaptive initialization for soft-prompt methods. To the best of our knowledge, this is the first study of applying meta-learning to prompting problem setting.

(2) We conduct extensive experiments on four different datasets with various few-shot settings, and results show the superiority of MetaPrompting over normally fine-tuned soft-prompt methods and SOTA meta-learning baselines.

(3) Further analysis experiments indicate that MetaPrompting significantly alleviates soft prompt initialization problem, and learns general meta-knowledge to counter the instability of prompt variance.
We also study MetaPrompting's compatibility with different meta-learning methods and give empirical analysis of their performance difference.

All code and data will be publicly available at \href{https://github.com/Dousia/MetaPrompting}{https://github.com/Dousia/MetaPrompting}.

\section{Preliminaries and Related Works}
In this section, we review related work and provide preliminaries about Language Model Prompting and Meta-learning. 
\subsection{Prompting Language Models}
\label{sec: 2.1}
Prompting methods are proposed to better apply pre-trained language models to downstream tasks by aligning them with pre-training tasks. For Masked Language Models (MLMs), the first step is to convert a sample text $x$ to $x_{prompt}$ by inserting prompt words which contain \texttt{[MASK]} tokens. Taking the news headline classification task as an example, the prompted text is given as:
\vspace{-1mm}
\begin{equation}
\label{eq:lm-classification}
\resizebox{.85\hsize}{!}{%
$x_{prompt} = \text{\texttt{[CLS]}~$x$~{The topic is}~\texttt{[MASK]}~. \texttt{[SEP]}}$,
}
\vspace{-1mm}
\end{equation}
where ``The topic is [MASK]'' are prompt tokens. Then, we ask pre-trained MLM to complete the prompted text $x_{prompt}$, and the word to be filled at \texttt{[MASK]} position is regarded as an answer. An answer-label map is then used to convert the word probability distribution at \texttt{[MASK]} to classification results. 
For example, answers `arts' and `culture' can be mapped to label `ARTS \& CULTURE', while `environment' can be mapped to label `ENVIRONMENT'. 
The average probability of each label's corresponding answers is computed as the label's final probability.

Early prompting methods, such as GPT-3~\cite{brown2020language} and PET/iPET~\cite{PET}, use hand-crafted prompt templates. Although promising results are achieved, the performance of these methods heavily relies on the selection of pre-defined prompt templates. Moreover, designing prompts is extremely time-consuming, and hand-crafted prompts may be sub-optimal. 



A number of recent works propose to automate the search of discrete prompt templates~\cite{shin2020autoprompt, gao2020making, davison2019commonsense, jiang2020can, haviv-etal-2021-bertese}, while others treat prompt tokens as continuous trainable parameters~\cite{li2021prefix, P-tuning, qin-eisner-2021-learning}. 
In this work, we follow P-tuning~\cite{P-tuning} to combine soft prompt and anchor tokens as templates. Instead of directly applying the model in few-shot tasks, however, we adopt meta-learning methods to find a better initialization point for both soft prompt embeddings and MLM parameters, because they are very sensitive to initialization in few-shot settings~\cite{li2021prefix, liu2021pre}.
Note that a recent work \cite{zhong2021adapting} also learns prompt model on a number of source domain tasks, but their method consumes heavy human labor to design hard prompts for each task, and directly fine-tunes the model without involving meta algorithms.

\subsection{Meta Learning}
Meta-learning algorithms can be classified into metric-based methods, model-based methods and optimization-based methods. 
Metric-based methods such as Siamese Network \cite{koch2015siamese}, Matching Network \cite{vinyals2016matching} and Prototypical Network \cite{PN}, are proposed to learn a metric space that gathers similar samples and separates distinct ones. 
Model-based meta-learning algorithms use additional meta learners to assist model prediction~\cite{graves2014neural, mishra2017simple, qiao2018few}.  

Different from above algorithms, optimization-based meta-learning methods learn to improve model's optimization procedure.
Optimization-based approach is more suitable for prompting models as it neither requires a specific task form (i.e., metric learning form) nor additional architecture (e.g. memory-augmented components in model-based algorithms).
\citet{Andrychowicz2016LearningTL} and \citet{Ravi2017OptimizationAA} train recurrent neural networks to transform vanilla gradient descent direction for better optimization results.  
MAML~\cite{MAML} optimizes model parameters to find a better initialization point, so that the model can adapt faster and better to unseen tasks. In addition to MAML, more elaborate methods also learn inner loop gradient descent direction~\cite{li2017meta} and inner step sizes~\cite{antoniou2019train}. Utilizing first-order derivatives, FOMAML~\cite{MAML} and Reptile \cite{Reptile} are proposed to reduce the memory consumption of high-order derivative calculation.



\section{Method}
Since prompt-based methods, especially those adopting soft prompts, are very sensitive to parameter initialization~\cite{li2021prefix, liu2021pre}, we introduce optimization-based meta-learning methods into prompting methods to find better initialization points for prompt-based models and further explore their capabilities in few-shot scenarios. 
In this section, we first introduce the prompt-based model tuning process used in our method (\S \ref{sec: 3.1}), and then describe how to construct Meta Prompting tasks (\S \ref{sec: 3.2}). 
Finally, we elaborate and formulate the Meta Prompting objective and parameter updating strategies (\S \ref{sec: 3.3} and \S \ref{sec: 3.4}). 

\begin{figure}[t]
 \centering
 \begin{tikzpicture}
 \draw (0,0 ) node[inner sep=0] {\includegraphics[width=0.9 \columnwidth, trim={1.6cm 15cm 7.5cm 1cm}, clip]{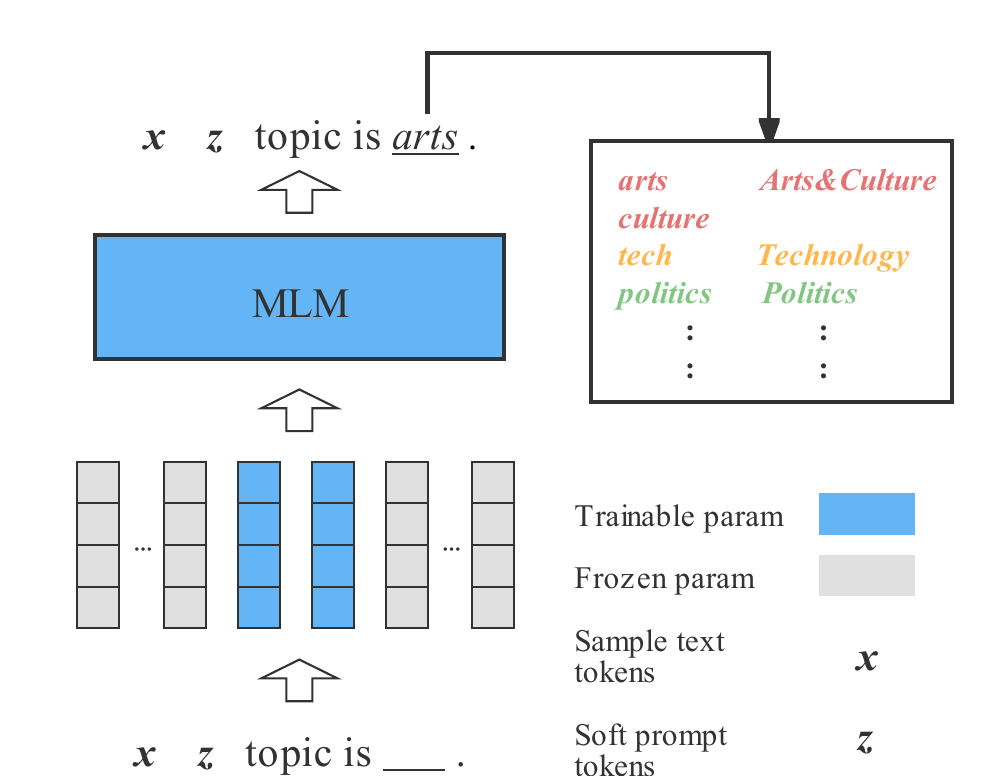}};
 \end{tikzpicture}
 \caption{
  An illustration of soft-prompting method. $x$ refers to the sample text, and $z$ represents soft prompt tokens. All trainable parameters are colored in blue, while fixed ones are colored in grey. 
 }\label{fig: model}
 \vspace*{-3mm}
\end{figure}

\subsection{Prompt-based Model Tuning}
\label{sec: 3.1}

In this work, we use soft prompts with anchor tokens. As illustrated in Figure \ref{fig: model}, prompt tokens consist of soft tokens which are represented as trainable parameters (blue) as well as anchor tokens which are fixed as the embeddings of specific words (grey). Hard-soft combined prompt templates render the model more flexible, while preserving enough semantic information to trigger the MLM to produce correct predictions.
Similar to P-tuning \cite{P-tuning}, we implement transformation layers on soft prompt embeddings, allowing them to escape from local minima smoothly. 


In this way, we define MLM parameters as $\theta$ and soft prompt token embeddings as $\phi$. Given a few-shot task $\tau$ where $\mathcal{D}_{\tau}=\{(\mathbf{x}_{i}, \mathbf{y}_{i})\}_{i\in\tau}$ represents training samples, the prompt tuning objective can be formulated as follows:
\vspace{-1mm}
\begin{equation}
\label{eq: adapt_obj}
\resizebox{.85\hsize}{!}{%
$\begin{split}
\theta^{*}, \phi^{*} &= \
\argmin\limits_{\theta, \phi}\mathcal{L}_{\mathcal{D}_{\tau}}(f_{\phi,\theta})\\
&=\argmax\limits_{\theta, \phi} \sum_{(\mathbf{x}_{i}, \mathbf{y}_{i})\in \mathcal{D}_{\tau}} \log P(\mathbf{y}_{i}|\mathbf{x}_{i}; \phi,\theta),
\end{split}$\quad
}
\vspace{-1mm}
\end{equation}
where $\mathcal{L}$ is the loss function, and $f_{\phi,\theta}$ is prompt-based model parameterized by MLM parameters $\theta$ and soft prompt embeddings $\phi$. 

$\mathcal{D}_{\tau}$ contains few labeled data because of the high annotation cost in real-world scenarios. As a result, the initialization of parameters $\theta$ and $\phi$ are more than crucial to the model's performance. 

\subsection{Constructing Meta Prompting Tasks}
\label{sec: 3.2}
To get a better initialization point for parameters $\theta$ and $\phi$, we propose to sample Meta Prompting tasks from accessible source data and conduct meta-training on these sampled tasks. This meta training process aims to simulate the model's adaptation to new few-shot tasks. 

We sample each Meta Prompting task $\tau_{i}$ as: 
\vspace{-0.8mm}
\begin{equation}
\label{}
\tau_{i}=(\mathcal{D}_{\tau_{i}}^{support}, \mathcal{D}_{\tau_{i}}^{query}),
\vspace{-0.8mm}
\end{equation}
where $\mathcal{D}_{\tau_{i}}^{support}$ indicates the support set and $\mathcal{D}_{\tau_{i}}^{query}$ indicates the query set in traditional few-shot learning settings. 
Note that meta training tasks and meta testing tasks should be sampled from different domains, to prevent the model from simply memorizing training samples.

\subsection{Applying Meta-learning to Prompting Models}
\label{sec: 3.3}
\begin{figure}[t]
 \centering
 \begin{tikzpicture}
 \draw (0,0 ) node[inner sep=0] {\includegraphics[width=0.97 \columnwidth, trim={1cm 10.5cm 8cm 1.3cm}, clip]{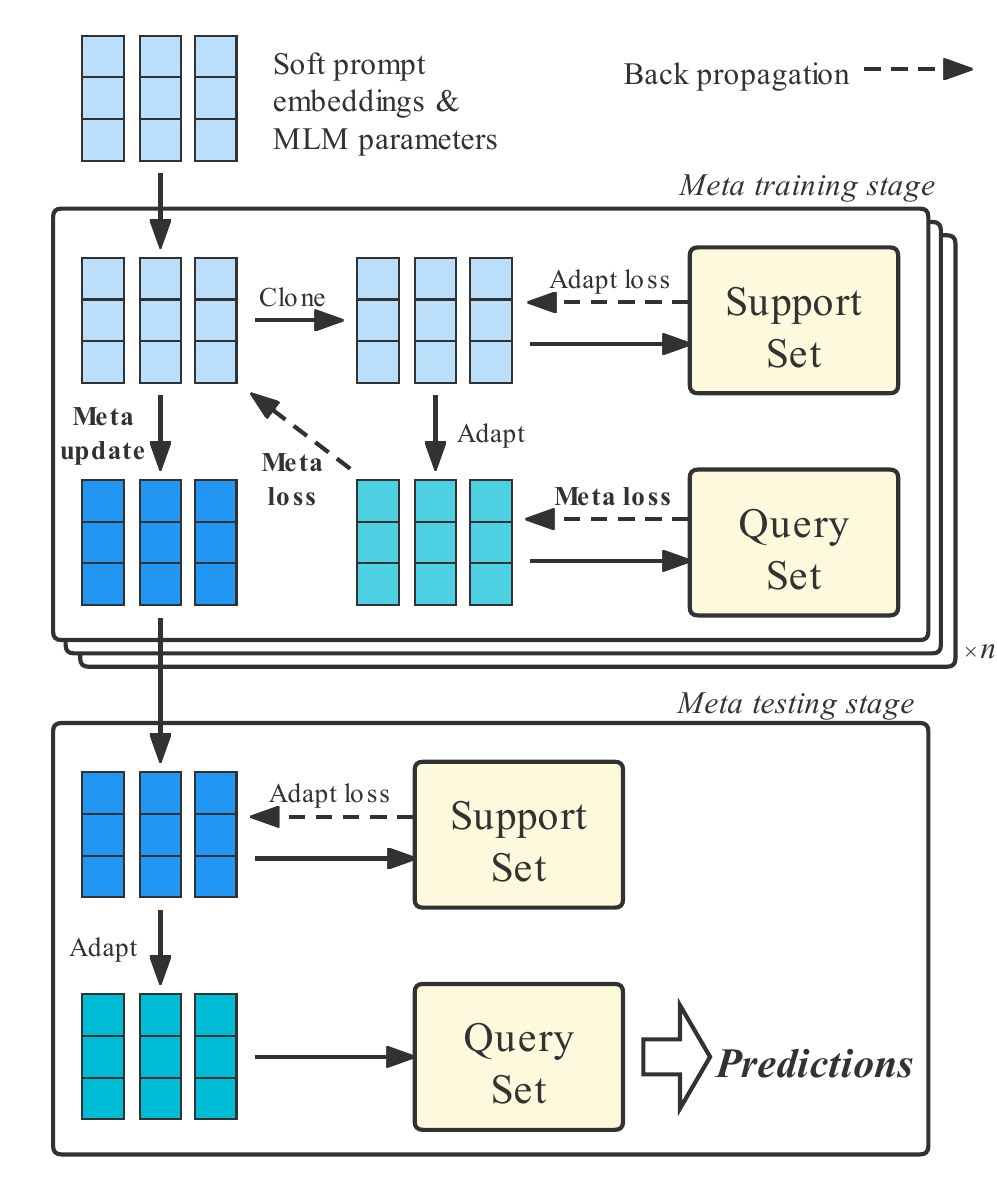}};
 \end{tikzpicture}
 \caption{
  Meta training and testing procedures of MetaPrompting.
 }\label{fig: meta}
 \vspace*{-4mm}
\end{figure}

After constructing Meta Prompting tasks, we train our prompting model on these tasks to find a better initialization point. 
Figure \ref{fig: meta} illustrates the meta training and meta testing procedures of MetaPrompting. 
Given a Meta Prompting task $\tau_{i}$, we clone the model's parameters and simulate the adaption process of few-shot tasks by updating cloned model parameters $\theta_{i}^0$ and $\phi_{i}^0$ on $\mathcal{D}_{\tau_{i}}^{support}$. 
The adaption objective is given in Equation \eqref{eq: adapt_obj}, and this process can be formulated as: 
\vspace{-1.2mm}
\begin{equation}
\label{eq: adapt}
\resizebox{.85\hsize}{!}{%
$\begin{split}
&\theta^{k}_{i} = \theta^{k-1}_{i} - \alpha\nabla_{\theta^{k-1}_{i}}\mathcal{L}_{\mathcal{D}^{support}_{\tau_i}}(f_{\phi^{k-1}_{i},\theta^{k-1}_{i}}), \\
&\phi^{k}_{i} = \phi^{k-1}_{i} - \alpha\nabla_{\phi^{k-1}_{i}}\mathcal{L}_{\mathcal{D}^{support}_{\tau_i}}(f_{\phi^{k-1}_{i},\theta^{k-1}_{i}}),
\end{split}$
}
\vspace{-1.2mm}
\end{equation}
where $\alpha$ indicates learning rate and $k=1, 2, 3, \dots$ indicates the inner step. 
The goal of learning with Meta Prompting tasks is to minimize the loss of the adapted prompting model, which is parameterized as $f_{\phi_{i},\theta_{i}}$, on $\mathcal{D}_{\tau_{i}}^{query}$. This objective can be described as follows:
\vspace{-1mm}
\begin{equation}
\label{}
\theta^{obj}, \phi^{obj} = \argmin\limits_{\theta, \phi} \mathcal{L}_{\mathcal{D}^{query}_{\tau_i}}(f_{\phi_{i},\theta_{i}}).
\vspace{-1mm}
\end{equation}


Optimizing towards this objective is to mimic real few-shot text classification scenarios, and enable prompting model to find a better initialization point for fast adaptation to new tasks. 
Let $\beta$ be the learning rate when updating model parameters on $\mathcal{D}_{\tau_{i}}^{query}$, and $\mathbf{H}$ be Hessian matrix. 
We formulate the second-order gradient of prompt parameter $\phi$  computed on $\mathcal{D}_{\tau_{i}}^{query}$ in the following form:
\vspace{-1.8mm}
\begin{equation}
\label{eq: MAML_update}
\begin{split}
\phi &\gets \phi - \beta \cdot g^{second}_{\phi}\\
&= \phi - \beta \nabla_{\phi} \mathcal{L}_{\mathcal{D}^{query}_{\tau_i}}(f_{\phi_{i},\theta_{i}})\\
&= \phi - \beta \nabla_{\phi_{i}} \mathcal{L}_{\mathcal{D}^{query}_{\tau_i}}(f_{\phi_{i},\theta_{i}}) \cdot \nabla_{\phi}(\phi_{i})\\
&= \phi - \beta \nabla_{\phi_{i}}\mathcal{L}_{\mathcal{D}^{query}_{\tau_i}}(f_{\phi_{i},\theta_{i}}) \cdot \\
&\qquad\quad(\mathbf{I} - \alpha\mathbf{H}_{\phi}(\mathcal{L}_{\mathcal{D}^{support}_{\tau_i}}(f_{\phi,\theta}))),\\
\end{split}
\vspace{-4.5mm}
\end{equation}
where we assume $\phi_{i}$ is $\phi$ adapted for one inner step on $\mathcal{D}^{support}_{\tau_i}$. In practice, inner steps can be increased for better performance. Pre-trained MLM parameters $\theta$ is updated in the same way as prompt parameters $\phi$ in Equation \eqref{eq: MAML_update}.



\subsection{Stable and Memory-efficient Meta Prompt Learning}
\label{sec: 3.4}



Although broadly used in meta-learning tasks, MAML suffers from training instability and exploding memory consumption when model size and inner step grow. To address the first problem, we follow \citet{antoniou2019train} to introduce Multi-Step Loss Back-propagation (MSLB) into prompting model tuning process.
In this way, prompting model parameters receive optimization information from each inner step during adaptation, alleviating the vanishing/exploding gradient problem in the stacked deep neural architecture constructed in adaptation process. 

As for the exploding memory consumption issue, we also explore to combine memory-efficient alternatives such as FOMAML~\cite{MAML} and Reptile~\cite{Reptile} with prompting model. FOMAML removes the high-order derivatives term in Equation~\eqref{eq: MAML_update}, providing a cheap approximation for MAML. 
Reptile updates model parameters towards the optimal point of each task, which is obtained by adapting the model on the support set samples. 
Equipped with these algorithms, MetaPrompting can learn meta knowledge with limited memory consumption.

\section{Experiment}
We conduct experiments by evaluating the proposed methods on four widely-used benchmark datasets with various low resource settings.

\subsection{Dataset}
Following \citet{Few-shotTC,Frog-GNN}, we use the following four text classification datasets for experiments, which provide well-founded benchmarks for the meta-train \& meta-test setting and vary in domain and text length.\footnote{For other datasets used in \citet{Few-shotTC}, RCV1~\cite{lewis2004rcv1} is not included due to overly long text lengths, while FewRel~\cite{han2018fewrel} is excluded because it does not provide each label's semantic meanings.}

\textbf{HuffPost headlines} contains around $36,900$ news headlines from $2012$ to $2018$ obtained from HuffPost~\cite{misra2018news,misra2021sculpting}. These headlines cover $41$ news categories and the average text length is $11$.

\textbf{Amazon product data} contains around $24,000$ product reviews from $1996$ to $2014$ from Amazon~\cite{Amazon}. These reviews contain $24$ categories corresponding to their respective product categories with varying text lengths. The average text length is $140$.

\textbf{20 Newsgroups} contains $18,820$ newsgroup documents of $20$ different topics~\cite{20News}. 
We used $20$news-$18828$ version following~\citet{Few-shotTC}. The average text length is $340$.

\textbf{Reuters} contains $620$ document-level news articles of $31$ different topics from 1987~\cite{lewis1997reuters}. 
The average text length is $168$.

We adopt the same pre-processing and data-splitting strategy with \citet{Few-shotTC} to process the above datasets. 


\subsection{Implementation}
We use the pre-trained BERT (bert-base-uncased) with HuggingFaces codebase~\cite{wolf2019huggingface} as the pre-trained language model.

For soft prompting model, we follow \citet{P-tuning} to use a two-layer biLSTM and a two-layer MLP to transform soft-prompt embeddings. 
We divide the learnable parameters of prompting model into two parts: pre-trained model and prompt embeddings. 
AdamW~\cite{AdamW} is used to optimize two types of parameters, with initial learning rates of $1$e$-5$ and $5$e$-5$, respectively. For pre-trained model parameters, we set weight decay to $0.1$. We also adopt linear warmup and linear decay strategy for learning rates.
Batch size is set as $16$ for all stages, and the model adapts for 15 epochs on test episodes.
We run 3 independent runs with random seeds for each setting.  

Before meta-training stage, we generate $10,000$ training episodes, $2,500$ validation episodes and $1,000$ testing episodes comprehensively and randomly. During the training stage, we train the model with $100$ sampled training episodes per epoch. When there is no validation accuracy increase for $10$ epochs, we apply early stopping. For meta-testing, we test the model on all $1,000$ test episodes and report the average accuracy. 

\begin{table*}[t]
\small
\resizebox{\textwidth}{36.5mm}{\begin{tabular}{l cc cc cc cc cc}
\toprule
\multicolumn{1}{c}{Method} &
\multicolumn{2}{c}{20 News} &
\multicolumn{2}{c}{Amazon} &
\multicolumn{2}{c}{HuffPost} &
\multicolumn{2}{c}{Reuters} &
\multicolumn{2}{c}{Average}
\\
\cmidrule(lr{0.5em}){2-3}\cmidrule(lr{0.5em}){4-5}\cmidrule(lr{0.5em}){6-7}\cmidrule(lr{0.5em}){8-9}\cmidrule(lr{0.5em}){10-11}
 & 1 shot & 5 shot & 1 shot & 5 shot & 1 shot & 5 shot & 1 shot & 5 shot & 1 shot & 5 shot 
\\
\midrule
\textsc{1-NN}       & $38.8$ &  $51.9$ & $51.4$ &  $67.1$ & $31.5$ &  $42.3$ & $ 57.8 $ &  $ 82.9 $ & $ 44.88 $ &  $ 61.05 $\\
\textsc{FT}        & $33.0$ &  $47.1$ & $45.7$ &  $63.9$ & $32.4$ &  $44.1$ & $ 70.9 $ &  $ 91.0 $ & $ 45.50 $ &  $ 61.53 $\\
\textsc{Proto}     & $37.8$ &  $46.5$ & $41.9$ &  $59.2$ & $34.8$ &  $50.2$ & $ 61.0 $ &  $ 72.1 $  & $43.88$ &  $57.00$\\
\textsc{MAML}      & $37.2$ & $48.6$ & $43.6$ & $62.4$ & $38.9$ & $53.7$ & $ 61.5 $ & $72.0$ & $45.30$ & $59.18$ \\
\textsc{RR}        & $44.8$ &  $64.3$ & $60.2$ &  $79.7$ & $37.6$ &  $59.5$ & $69.1$ &  $93.0$ & $52.93$ &  $74.13$\\
\textsc{DS}~\citeyearpar{Few-shotTC} &  $52.1$ & $68.3$ & $62.6$ & $81.1$ & $43.0$ & $63.5$ & $81.8$ & $96.0$ & $59.88$ & $77.23$\\
\textsc{DE-MLMAN}~\citeyearpar{DE}        & $-$ &  $-$ & $-$ & $- $ & $49.7$ &  $60.9$ & $-$ &  $-$ & $-$ &  $-$\\
\textsc{DE-MAML}~\citeyearpar{DE}         & $-$ &  $-$ & $-$ & $- $ & $51.8$ &  $67.3$ & $-$ &  $-$ & $-$ &  $-$\\
\textsc{DE-Proto}~\citeyearpar{DE}        & $-$ &  $-$ & $-$ & $- $ & $52.3$ &  $69.6$ & $-$ &  $-$ & $-$ &  $-$\\
\textsc{KGML-Proto}~\citeyearpar{KGML}        & $-$ &  $-$ & $58.6$ &  $74.5$ & $42.3$ &  $58.7$ & $-$ &  $-$ & $-$ &  $-$\\
\textsc{KGML-MAML}~\citeyearpar{KGML}        & $-$ &  $-$ & $51.4$ &  $58.8$ & $44.2$ &  $54.1$ & $-$ &  $-$ & $-$ &  $-$\\
\textsc{P-Tuning}~\citeyearpar{P-tuning}        & $61.20$ &  $71.47$ & $62.18$ &  $79.13$ & $54.48$ &  $65.75$ & $90.01$ &  $96.66$ & $66.97$ &  $78.25$\\
\textsc{Frog-GNN}~\citeyearpar{Frog-GNN}        & $-$ &  $-$ & $71.5$ &  $83.6$ & $54.1$ &  $69.6$ & $-$ &  $-$ & $-$ &  $-$\\
\textsc{LaSAML-PN}~\citeyearpar{LaSAML}         & $-$ &  $-$ & $-$ & $- $ & $62.1$ &  $70.1$ & $-$ &  $-$ & $-$ &  $-$\\
\textsc{ContrastNet}~\citeyearpar{chen2022contrastnet}         & $71.74$ &  $\bm{81.57}$ & $76.13$ & $85.17$ & $53.06$ &  $65.32$ & $86.42$ &  $95.33$ & $71.84$ &  $81.85$\\

\midrule
\textsc{Ours~(Pretrain Init)} & $72.52$  & $76.32$ & $75.12$ & $83.27$ & $70.82$ & $75.47$ & $95.07$ & $\bm{97.29}$ & $78.38$ & $83.09$\\
\textsc{Ours~(Meta Init)} & $\bm{73.75}$  & ${76.57}$ & $\bm{77.65}$ & $\bm{85.54}$ & $\bm{71.93}$ & $\bm{76.32}$ & $\bm{95.20}$ & $97.17$ & $\bm{79.63}$ & $\bm{83.90}$\\
\bottomrule
\end{tabular}}
\centering
\caption{Results of 1-shot and 5-shot classification on four datasets in terms of accuracy.
The rows below the mid-line are results of MetaPrompting. ‘-’ means that the result of this dataset is not given in the original paper. We do not show standard deviation of our experiment results here due to space limits. Full results can be found in Appendix~\ref{appendix: A}.}
\label{tbl:main}
\vspace*{-3mm}
\end{table*}

\subsection{Baselines}
We compare with the following baselines:

\textbf{1-NN} is a $1$-nearest-neighbor classifier based on Euclidean distance.

\textbf{FT}~\cite{FT} pre-trains a classifier on source domain data, and then fine-tunes (FT) it on each support set before evaluation.

\textbf{RR}~\cite{RR} adopts ridge regression (RR) for classification.

\textbf{MAML}~\cite{MAML} meta-learns a classifier with MAML algorithm, so that the model can adapt faster and better to target domain tasks. 

\textbf{Prototypical network}~\cite{PN} is a metric-based method which meta-learns a metric space by minimizing the Euclidean distance between the centroid of each category and the corresponding samples. 

\textbf{DS}~\cite{Few-shotTC} is trained within a meta-learning framework to map the distribution signatures (DS), i.e., characteristics of the underlying word distributions, into attention scores to extract more transferable features.\footnote{The above 6 baselines uses fastText embeddings~\cite{joulin2016fasttext} and each word's inverse document frequency to produce sentence embeddings. }

\textbf{DE}~\cite{DE} generates distinct label representations that embed information specific to each label to aid classification tasks. During experiments, it is combined with MAML (DE-MAML) and prototypical network (DE-PROTO), as well as MLMAN~\cite{MLMAN} (DE-MLMAN).

\textbf{KGML}~\cite{KGML} extracts additional representation for each sentence from external knowledge base, to bridge the gap between meta-training and meta-testing tasks. During experiments, it works with MAML (KGML-MAML) and prototypical network (KGML-Proto).

\textbf{P-tuning}~\cite{P-tuning} is a prompt-based method that uses masked language model to convert target tasks into cloze problems. It employs soft-prompting techniques to optimize prompts in continuous space.

\textbf{Frog-GNN}~\cite{Frog-GNN} is a graph neural network based method, which extracts better query representations with multi-perspective aggregation of graph node neighbors.

\textbf{LaSAML-PN}~\cite{LaSAML} is a meta-learning framework that mines semantic information in labels and attaches it to the sentence as the input of the encoder to obtain discriminative sentence embeddings. 

\textbf{ContrastNet}~\cite{chen2022contrastnet} is the SOTA method. It introduces instance-level and task-level regularization loss into vanilla contrastive learning model based on BERT representations for better generalization performance. The regularization loss is computed with samples augmented by an additional BERT model.

\subsection{Main Results}
\label{sec: 4.4}
We evaluate the proposed methods in both 5-way 1-shot and 5-way 5-shot settings and report performance on four different datasets with different text styles. 
As shown in Table \ref{tbl:main}, our model outperforms previous SOTA method ContrastNet without using additional PLM. 
Averagely, our model improves 1-shot accuracy by $7.79$ ($10.84\%\uparrow$) and 5-shot accuracy by $2.05$ ($2.50\%\uparrow$) across four datasets. 
MetaPrompting gains less improvement on 20Newsgroup and Amazon, because their labels are hard to interpret as natural words, which poses difficulties for prompting models~\cite{shin2020autoprompt,cui2022prototypical}. 
Various methods are proposed to address this problem~\cite{shin2020autoprompt,gao2020making,hambardzumyan2021warp,cui2022prototypical,jiang2021can}. In this work, however, we do not spend much effort on elaborate answer design but instead focus on soft prompt initialization problem. Although equipped with simply designed answer sets, MetaPrompting still achieves new state-of-the-art performance across the four datasets.



Meanwhile, we have following observations based on Table \ref{tbl:main}:

\textbf{(1)} Compared with other soft-prompting methods, i.e., P-tuning, our method obtains superior results. Although meta learning is conducted on completely different source domain Meta Prompting tasks, our method still learns a better prompt model initialization point, which allows faster and better adaption to new prompting tasks. 

\textbf{(2)} When compared to traditional supervised learning methods, such as FT, all prompt-based methods achieve significant improvements, which demonstrates the effectiveness of prompting mechanism in narrowing the gap between pretraining and downstream tasks. 

\textbf{(3)} Metric learning-based baselines, such as ContrastNet and LASAML-PN, perform as the strongest baselines on Amazon and HuffPost datasets, respectively.
We find that directly using prompt-based method may not necessarily perform better, because of the absence of domain-related initialization. 
The proposed MetaPrompting alleviates the above issue and achieves new state-of-the-art performance. 
Among strong metric-learning baselines, Frog-GNN conducts transductive learning with additional label propagation information, and ContrastNet uses an additional BERT model to regularize the main model with augmented data. Our model achieves better performance without implementing any of above tricks. 

\textbf{(4)} Compared with other optimization-based meta-learning methods such as MAML, DE-MAML and KGML-MAML, MetaPrompting consistently performs better, demonstrating good compatibility between prompting methods and meta-learning. Note that KGML-MAML adopts an additional knowledge base, while our model does not but still achieves better performance.

\textbf{(5)} To further demonstrate meta learning method's effectiveness in prompt learning, we conduct ablations study by removing the Meta Prompting Objective and learn an initialization by pre-training soft prompt model on Meta Prompting Tasks described in Section \ref{sec: 3.2}. The results are shown as \textsc{OURs (Pretrain Init)}. 
Performance drops are witnessed across all four datasets and few-shot settings, validating the necessity of meta objectives in finding a better initialization.




\subsection{Analysis}
In this part, we analyze the proposed method from different aspects.


\paragraph{MetaPrompting tackles soft prompt initialization problem.}
Main results displayed in Section \ref{sec: 4.4} are obtained with LM parameters tuned for fair comparison with previous SOTA baselines. 
To further validate the importance of learning a good prompt initialization, we freeze PLM's parameters while leaving soft prompt parameters unfrozen to only learn a better prompt initialization on source domains. We test our meta-learning-based initialization strategy against random initialization, and the results are shown in Table \ref{tbl:initialization}. The randomly initialized soft prompt baseline performs poorly and unstably, while our method consistently yields better results with lower variance across four datasets, which verifies our hypothesis and the validity of the MetaPrompting.



\begin{table}[t]
\small
\begin{tabular}{c cc cc cc cc}
\toprule
Method & HuffPost & Amazon & 20 News & Reuters
\\
\midrule
\textsc{Baseline}       & $65.75$ & $79.13$ & $71.47$ & $96.66$\\
\textsc{Ours}       &  $73.06$ & $83.64$ & $75.60$ & $97.26$\\
\bottomrule
\end{tabular}
\centering
\caption{
PLM frozen, MetaPrompting still achieves better performance over randomly initialized baseline on test domains. Results are given in 5 shot setting. 
}\label{tbl:initialization}
\end{table}

\begin{table}[t]
\small
\begin{tabular}{c c c}
\toprule
Source domain & Target domain & Acc
\\
\midrule
$-$ & HuffPost & $65.75$\\
Metatuning & HuffPost & $67.46$\\
Reuters & HuffPost & $69.18$\\
20 Newsgroup & HuffPost & $71.04$\\
Amazon & HuffPost & $71.47$\\
HuffPost (Diff. label set) & HuffPost & $76.32$\\
\bottomrule
\end{tabular}
\centering
\caption{
Given irrelevant source domain data, MetaPrompting still learns meta knowledge to improve the performance on target domains. 
}\label{tbl:OOD}
\end{table}

\paragraph{MetaPrompting learns general meta-knowledge from various source domains.}
Although MetaPrompting shows promising results when meta trained on source domain tasks from the same dataset, it is impractical to always build corresponding meta-training tasks for each few-shot scenario. To this end, we conduct meta-training on Out-Of-Domain (OOD) tasks to better analyze MetaPrompting's ability in transferring meta-knowledge from various source domains.

Table \ref{tbl:OOD} shows the results of 5-shots setting. Even given irrelevant meta-training data and prompt templates from other datasets, MetaPrompting still learns meta knowledge to tackle target domain tasks and outperforms the baseline robustly. Among OOD datasets, Metatuning~\cite{zhong2021adapting} contains a series of text classification tasks, and each task is accompanied by several hand-crafted questions which require yes/no answers. The task formulation of Metatuning is distinct from HuffPost. However, MetaPrompting still makes it to transfer meta-knowledge from Metatuning to HuffPost's target domains, improving model performance by approximately 2 points. Although MetaPrompting's performance varies among source domain tasks according to their data quality for generalization purposes, the proposed model outperforms the baseline across all source domain tasks, verifying MetaPrompting's effectiveness in transferring meta-knowledge.

\paragraph{Anti-disturbance analysis}
We expect the meta-learned initialization alleviates prompting models' susceptibility to varying prompt forms.
To verify this, we test the prompting model with multiple different prompt forms and report the standard deviation. 
Specifically, we add two more discrete prompt templates, and randomly replace the template tokens with pseudo tokens to test MetaPrompting's robustness across different templates.\footnote{We add ``\textit{The \underline{topic/product category}: [MASK]. Input: \underline{$x$}}'' and ``\textit{\underline{$x$}. What is the \underline{topic/product category} ? [MASK].}'', where topic and product category are used for HuffPost and Amazon dataset respectively.}

Table \ref{tbl:anti} shows the results. 
While changing the prompting form indeed impacts the performance for both our method and normal soft prompting methods, the proposed meta-learning method significantly reduces performance fluctuation, showing impressive anti-disturbance ability. 
Therefore, the proposed method is promising in real-world applications, because prompt designing requires heavy workload and domain-specific knowledge. Applying MetaPrompting significantly reduces the cost of prompt engineering. 

\begin{table}[t]
\small
\begin{tabular}{c cc cc cc cc}
\toprule
\multirow{2}{*}{Method} &
\multicolumn{2}{c}{HuffPost} &
\multicolumn{2}{c}{Amazon} 
\\
& 1 shot & 5 shot & 1 shot & 5 shot
\\
\midrule
\textsc{P-tuning}       &  $\pm{3.46}$ &  $\pm{1.90}$ & $\pm{5.30}$ & $\pm{1.85}$\\
\textsc{OURs}       & $\pm{0.23}$ & $\pm{0.09}$ &$\pm{0.17}$ & $\pm{0.45}$\\
\bottomrule
\end{tabular}
\centering
\caption{
Analysis for anti-disturbance against changing of prompting form.
}\label{tbl:anti}
\end{table}

\paragraph{Applying different meta-learning methods to prompting models.}
In this part, we conduct empirical analysis on different optimization-based meta-learning methods applied in prompting models. Results are shown in Table~\ref{tbl:meta}. Stabilizing MAML training procedure, MAML++ performs the best among all methods, while Reptile fails to achieve comparable performance with others.\footnote{We only include the MSLB trick of MAML++ \cite{antoniou2019train} due to the incompatibility (BN layer tricks) or trivial performance improvement (Per-step adaption loss, cosine annealing learning rates).} We attribute Reptile's low performance to PLM's sensitivity to parameter tuning process, which can be distorted by Reptile's parameter updating strategy. MAML and FOMAML show similar results, because MetaPrompting's slow tuning process 
narrows the gap between their calculated gradient during meta-training.

\begin{table}[t]
\small
\resizebox{\linewidth}{8.5mm}{\begin{tabular}{c c c c c}
\toprule
Setting & MAML++ & MAML & FOMAML & Reptile
\\
\midrule
\textsc{1 shot}       & $71.93$ & $71.43$ & $70.56$ & $69.76$\\
\textsc{5 shot}       & $76.32$ & $76.04$ & $76.08$ & $74.09$\\
\bottomrule
\end{tabular}}
\centering
\caption{
MetaPrompting's performance with different meta learning methods. 
}\label{tbl:meta}
\vspace*{-2mm}
\end{table}

\paragraph{Analysis for learning procedure of prompting methods.}
We analyze the decreasing trend of adaptation loss to better understand the learning procedure of soft-prompt model.
Specifically, we visualize model adaptation loss curve during meta-testing on 5 shot Amazon dataset.

\begin{figure}[t]
 \centering
 
 \begin{tikzpicture}
 \draw (0,0 ) node[inner sep=0] {\includegraphics[width=0.95 \columnwidth, trim={0.2cm 0.1cm 1cm 1.1cm}, clip]{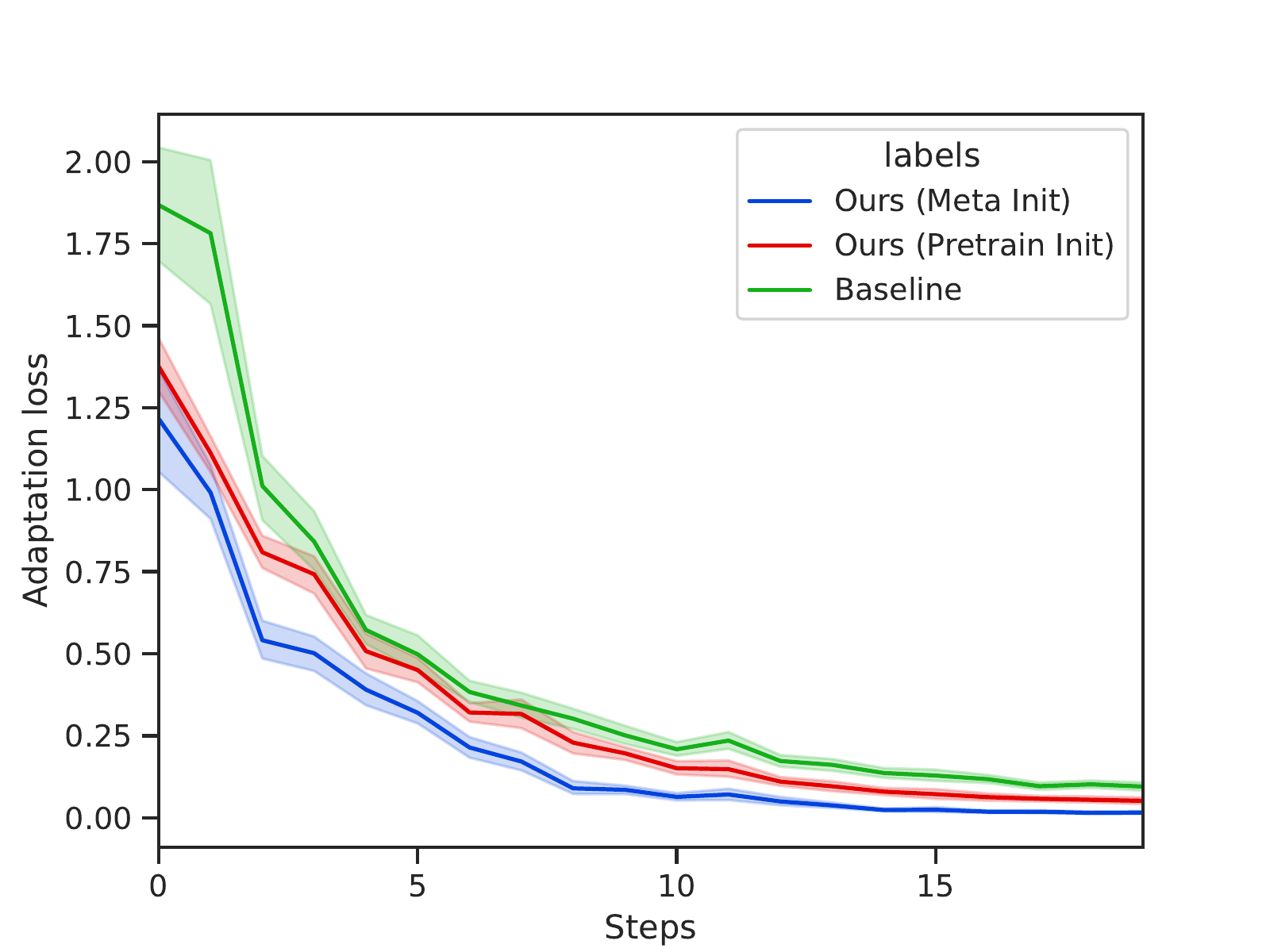}};
 \end{tikzpicture}
 \caption{
 Analysis on prompt learning process.
 }\label{fig:loss}
\end{figure}

As shown in Figure \ref{fig:loss}, task-related initialization (Ours (Pretrain Init)) helps the model converge faster and end up at a lower position than randomly initialized baseline.
The proposed meta-learning-based method (Ours (Meta Init)) further improves the learning process in new tasks, indicating that the meta-learned initialization point contains more generalizable meta knowledge to aid new tasks.

\section{Conclusion}
In this paper, we introduce a generalized optimization-based meta-learning
approach MetaPrompting for few-shot NLP problems. 
Utilizing sampled meta tasks and meta-learning-based optimization, MetaPrompting learns to find an initialization that alleviates soft prompt initialization problem, and allows better and faster adaption to new tasks. 
Extensive experiments on four few-shot learning benchmarks show that MetaPrompting significantly outperforms vanilla soft-prompting models and strong meta-learning baselines, achieving new state-of-the-art performance.

\section{Acknowledgement}

This work was supported by the National Key R\&D Program of China via grant 2020AAA0106501 and the National Natural Science Foundation of China (NSFC) via grant 61976072 and 62176078.

\bibliography{ref}

\begin{thebibliography}{52}
\expandafter\ifx\csname natexlab\endcsname\relax\def\natexlab#1{#1}\fi

\bibitem[{Andrychowicz et~al.(2016)Andrychowicz, Denil, Colmenarejo, Hoffman,
  Pfau, Schaul, and de~Freitas}]{Andrychowicz2016LearningTL}
Marcin Andrychowicz, Misha Denil, Sergio~Gomez Colmenarejo, Matthew~W. Hoffman,
  David Pfau, Tom Schaul, and Nando de~Freitas. 2016.
\newblock Learning to learn by gradient descent by gradient descent.
\newblock In \emph{Proc. of NIPS}.

\bibitem[{Antoniou et~al.(2019)Antoniou, Edwards, and
  Storkey}]{antoniou2019train}
Antreas Antoniou, Harri Edwards, and Amos Storkey. 2019.
\newblock How to train your maml.
\newblock In \emph{Proc. of ICLR}.

\bibitem[{Bao et~al.(2019)Bao, Wu, Chang, and Barzilay}]{Few-shotTC}
Yujia Bao, Menghua Wu, Shiyu Chang, and Regina Barzilay. 2019.
\newblock Few-shot text classification with distributional signatures.
\newblock In \emph{Proc. of {ICLR}}.

\bibitem[{Bertinetto et~al.(2019)Bertinetto, Henriques, Torr, and Vedaldi}]{RR}
L~Bertinetto, J~Henriques, PHS Torr, and A~Vedaldi. 2019.
\newblock Meta-learning with differentiable closed-form solvers.
\newblock In \emph{Proc. of ICLR}.

\bibitem[{Brown et~al.(2020)Brown, Mann, Ryder, Subbiah, Kaplan, Dhariwal,
  Neelakantan, Shyam, Sastry, Askell et~al.}]{brown2020language}
Tom Brown, Benjamin Mann, Nick Ryder, Melanie Subbiah, Jared~D Kaplan, Prafulla
  Dhariwal, Arvind Neelakantan, Pranav Shyam, Girish Sastry, Amanda Askell,
  et~al. 2020.
\newblock Language models are few-shot learners.
\newblock In \emph{Proc. of NIPS}, volume~33, pages 1877--1901.

\bibitem[{Chen et~al.(2022)Chen, Zhang, Mao, and Xue}]{chen2022contrastnet}
Junfan Chen, Richong Zhang, Yongyi Mao, and Jie Xue. 2022.
\newblock Contrastnet: A contrastive learning framework for few-shot text
  classification.
\newblock In \emph{Proc. of AAAI}.

\bibitem[{Chen et~al.(2019)Chen, Liu, Kira, Wang, and Huang}]{FT}
Wei-Yu Chen, Yen-Cheng Liu, Zsolt Kira, Yu-Chiang~Frank Wang, and Jia-Bin
  Huang. 2019.
\newblock A closer look at few-shot classification.
\newblock In \emph{Proc. of {ICLR}}.

\bibitem[{Cui et~al.(2022)Cui, Hu, Ding, Huang, and Liu}]{cui2022prototypical}
Ganqu Cui, Shengding Hu, Ning Ding, Longtao Huang, and Zhiyuan Liu. 2022.
\newblock Prototypical verbalizer for prompt-based few-shot tuning.
\newblock In \emph{Proceedings of the 60th Annual Meeting of the Association
  for Computational Linguistics (Volume 1: Long Papers)}, pages 7014--7024.

\bibitem[{Davison et~al.(2019)Davison, Feldman, and
  Rush}]{davison2019commonsense}
Joe Davison, Joshua Feldman, and Alexander~M Rush. 2019.
\newblock Commonsense knowledge mining from pretrained models.
\newblock In \emph{Proc. of {EMNLP-IJCNLP}}, pages 1173--1178.

\bibitem[{Finn et~al.(2017)Finn, Abbeel, and Levine}]{MAML}
Chelsea Finn, Pieter Abbeel, and Sergey Levine. 2017.
\newblock Model-agnostic meta-learning for fast adaptation of deep networks.
\newblock In \emph{Proc. of {ICML}}, pages 1126--1135.

\bibitem[{Gao et~al.(2021)Gao, Fisch, and Chen}]{gao2020making}
Tianyu Gao, Adam Fisch, and Danqi Chen. 2021.
\newblock Making pre-trained language models better few-shot learners.
\newblock In \emph{Proc. of ACL-IJCNLP}, pages 3816--3830.

\bibitem[{Graves et~al.(2014)Graves, Wayne, and Danihelka}]{graves2014neural}
Alex Graves, Greg Wayne, and Ivo Danihelka. 2014.
\newblock Neural turing machines.
\newblock \emph{arXiv preprint arXiv:1410.5401}.

\bibitem[{Hambardzumyan et~al.(2021)Hambardzumyan, Khachatrian, and
  May}]{hambardzumyan2021warp}
Karen Hambardzumyan, Hrant Khachatrian, and Jonathan May. 2021.
\newblock Warp: Word-level adversarial reprogramming.
\newblock In \emph{Proceedings of the 59th Annual Meeting of the Association
  for Computational Linguistics and the 11th International Joint Conference on
  Natural Language Processing (Volume 1: Long Papers)}, pages 4921--4933.

\bibitem[{Han et~al.(2018)Han, Zhu, Yu, Wang, Yao, Liu, and
  Sun}]{han2018fewrel}
Xu~Han, Hao Zhu, Pengfei Yu, Ziyun Wang, Yuan Yao, Zhiyuan Liu, and Maosong
  Sun. 2018.
\newblock Fewrel: A large-scale supervised few-shot relation classification
  dataset with state-of-the-art evaluation.
\newblock In \emph{Proc. of EMNLP}, pages 4803--4809.

\bibitem[{Haviv et~al.(2021)Haviv, Berant, and
  Globerson}]{haviv-etal-2021-bertese}
Adi Haviv, Jonathan Berant, and Amir Globerson. 2021.
\newblock {BERT}ese: Learning to speak to {BERT}.
\newblock In \emph{Proc. of EACL}, pages 3618--3623.

\bibitem[{He and McAuley(2016)}]{Amazon}
Ruining He and Julian McAuley. 2016.
\newblock Ups and downs: Modeling the visual evolution of fashion trends with
  one-class collaborative filtering.
\newblock In \emph{proc. of {WWW}}, pages 507--517.

\bibitem[{Jiang et~al.(2021)Jiang, Araki, Ding, and Neubig}]{jiang2021can}
Zhengbao Jiang, Jun Araki, Haibo Ding, and Graham Neubig. 2021.
\newblock How can we know when language models know? on the calibration of
  language models for question answering.
\newblock \emph{Transactions of the Association for Computational Linguistics},
  9:962--977.

\bibitem[{Jiang et~al.(2020)Jiang, Xu, Araki, and Neubig}]{jiang2020can}
Zhengbao Jiang, Frank~F Xu, Jun Araki, and Graham Neubig. 2020.
\newblock How can we know what language models know?
\newblock \emph{Transactions of the Association for Computational Linguistics},
  8:423--438.

\bibitem[{Joulin et~al.(2016)Joulin, Grave, Bojanowski, Douze, J{\'e}gou, and
  Mikolov}]{joulin2016fasttext}
Armand Joulin, Edouard Grave, Piotr Bojanowski, Matthijs Douze, H{\'e}rve
  J{\'e}gou, and Tomas Mikolov. 2016.
\newblock Fasttext. zip: Compressing text classification models.
\newblock \emph{arXiv preprint arXiv:1612.03651}.

\bibitem[{Koch et~al.(2015)Koch, Zemel, Salakhutdinov et~al.}]{koch2015siamese}
Gregory Koch, Richard Zemel, Ruslan Salakhutdinov, et~al. 2015.
\newblock Siamese neural networks for one-shot image recognition.
\newblock In \emph{ICML deep learning workshop}, volume~2.

\bibitem[{Lang(1995)}]{20News}
Ken Lang. 1995.
\newblock Newsweeder: Learning to filter netnews.
\newblock In \emph{Proc. of {ICML}}, pages 331--339.

\bibitem[{Lester et~al.(2021)Lester, Al{-}Rfou, and Constant}]{lester2021power}
Brian Lester, Rami Al{-}Rfou, and Noah Constant. 2021.
\newblock The power of scale for parameter-efficient prompt tuning.
\newblock In \emph{Proc. of {EMNLP}}, pages 3045--3059.

\bibitem[{Lewis(1997)}]{lewis1997reuters}
D~Lewis. 1997.
\newblock Reuters-21578 text categorization test collection.
\newblock \emph{Distribution 1.0, AT\&T Labs-Research}.

\bibitem[{Lewis et~al.(2004)Lewis, Yang, Russell-Rose, and Li}]{lewis2004rcv1}
David~D Lewis, Yiming Yang, Tony Russell-Rose, and Fan Li. 2004.
\newblock Rcv1: A new benchmark collection for text categorization research.
\newblock \emph{Journal of machine learning research}, 5:361--397.

\bibitem[{Li and Liang(2021)}]{li2021prefix}
Xiang~Lisa Li and Percy Liang. 2021.
\newblock Prefix-tuning: Optimizing continuous prompts for generation.
\newblock In \emph{Proc. of ACL-IJCNLP}, pages 4582--4597.

\bibitem[{Li et~al.(2017)Li, Zhou, Chen, and Li}]{li2017meta}
Zhenguo Li, Fengwei Zhou, Fei Chen, and Hang Li. 2017.
\newblock Meta-sgd: Learning to learn quickly for few-shot learning.
\newblock \emph{arXiv preprint arXiv:1707.09835}.

\bibitem[{Liu et~al.(2021{\natexlab{a}})Liu, Shen, Zhang, Dolan, Carin, and
  Chen}]{liu2021makes}
Jiachang Liu, Dinghan Shen, Yizhe Zhang, Bill Dolan, Lawrence Carin, and Weizhu
  Chen. 2021{\natexlab{a}}.
\newblock What makes good in-context examples for gpt-$3 $?
\newblock \emph{arXiv preprint arXiv:2101.06804}.

\bibitem[{Liu et~al.(2021{\natexlab{b}})Liu, Yuan, Fu, Jiang, Hayashi, and
  Neubig}]{liu2021pre}
Pengfei Liu, Weizhe Yuan, Jinlan Fu, Zhengbao Jiang, Hiroaki Hayashi, and
  Graham Neubig. 2021{\natexlab{b}}.
\newblock Pre-train, prompt, and predict: A systematic survey of prompting
  methods in natural language processing.
\newblock \emph{arXiv preprint arXiv:2107.13586}.

\bibitem[{Liu et~al.(2021{\natexlab{c}})Liu, Zheng, Du, Ding, Qian, Yang, and
  Tang}]{P-tuning}
Xiao Liu, Yanan Zheng, Zhengxiao Du, Ming Ding, Yujie Qian, Zhilin Yang, and
  Jie Tang. 2021{\natexlab{c}}.
\newblock Gpt understands, too.
\newblock \emph{arXiv preprint arXiv:2103.10385}.

\bibitem[{Loshchilov and Hutter(2018)}]{AdamW}
Ilya Loshchilov and Frank Hutter. 2018.
\newblock Decoupled weight decay regularization.
\newblock In \emph{Proc. of {ICLR}}.

\bibitem[{Luo et~al.(2021)Luo, Liu, Lin, and Zhang}]{LaSAML}
Qiaoyang Luo, Lingqiao Liu, Yuhao Lin, and Wei Zhang. 2021.
\newblock Don’t miss the labels: Label-semantic augmented meta-learner for
  few-shot text classification.
\newblock In \emph{Findings of {ACL-IJCNLP}}, pages 2773--2782.

\bibitem[{Mishra et~al.(2018)Mishra, Rohaninejad, Chen, and
  Abbeel}]{mishra2017simple}
Nikhil Mishra, Mostafa Rohaninejad, Xi~Chen, and Pieter Abbeel. 2018.
\newblock A simple neural attentive meta-learner.
\newblock In \emph{Proc. of ICLR}.

\bibitem[{Misra(2018)}]{misra2018news}
Rishabh Misra. 2018.
\newblock News category dataset, 06.

\bibitem[{Misra and Grover(2021)}]{misra2021sculpting}
Rishabh Misra and Jigyasa Grover. 2021.
\newblock \emph{Sculpting Data for ML: The first act of Machine Learning}.

\bibitem[{Nichol et~al.(2018)Nichol, Achiam, and Schulman}]{Reptile}
Alex Nichol, Joshua Achiam, and John Schulman. 2018.
\newblock On first-order meta-learning algorithms.
\newblock \emph{arXiv preprint arXiv:1803.02999}.

\bibitem[{Ohashi et~al.(2021)Ohashi, Takayama, Kajiwara, and Arase}]{DE}
Sora Ohashi, Junya Takayama, Tomoyuki Kajiwara, and Yuki Arase. 2021.
\newblock Distinct label representations for few-shot text classification.
\newblock In \emph{Proc. of {ACL-IJCNLP}}, pages 831--836.

\bibitem[{Qiao et~al.(2018)Qiao, Liu, Shen, and Yuille}]{qiao2018few}
Siyuan Qiao, Chenxi Liu, Wei Shen, and Alan~L Yuille. 2018.
\newblock Few-shot image recognition by predicting parameters from activations.
\newblock In \emph{Proc. of {CVPR}}, pages 7229--7238.

\bibitem[{Qin and Eisner(2021)}]{qin-eisner-2021-learning}
Guanghui Qin and Jason Eisner. 2021.
\newblock Learning how to ask: Querying {LM}s with mixtures of soft prompts.
\newblock In \emph{Proc. of {NAACL-HLT}}, pages 5203--5212.

\bibitem[{Ravi and Larochelle(2017)}]{Ravi2017OptimizationAA}
Sachin Ravi and H.~Larochelle. 2017.
\newblock Optimization as a model for few-shot learning.
\newblock In \emph{Proc. of ICLR}.

\bibitem[{Schick et~al.(2020)Schick, Schmid, and
  Sch{\"{u}}tze}]{schick2020coling}
Timo Schick, Helmut Schmid, and Hinrich Sch{\"{u}}tze. 2020.
\newblock Automatically identifying words that can serve as labels for few-shot
  text classification.
\newblock In \emph{Proc. of {COLING}}, pages 5569--5578.

\bibitem[{Schick and Sch{\"{u}}tze(2021)}]{schick2021eacl}
Timo Schick and Hinrich Sch{\"{u}}tze. 2021.
\newblock Exploiting cloze-questions for few-shot text classification and
  natural language inference.
\newblock In \emph{Proc. of {EACL}}, pages 255--269.

\bibitem[{Schick and Sch{\"u}tze(2021)}]{PET}
Timo Schick and Hinrich Sch{\"u}tze. 2021.
\newblock It’s not just size that matters: Small language models are also
  few-shot learners.
\newblock In \emph{Proc. of NAACL}, pages 2339--2352.

\bibitem[{Shin et~al.(2020)Shin, Razeghi, Logan~IV, Wallace, and
  Singh}]{shin2020autoprompt}
Taylor Shin, Yasaman Razeghi, Robert~L Logan~IV, Eric Wallace, and Sameer
  Singh. 2020.
\newblock Autoprompt: Eliciting knowledge from language models with
  automatically generated prompts.
\newblock In \emph{Proc. of {EMNLP}}, pages 4222--4235.

\bibitem[{Snell et~al.(2017)Snell, Swersky, and Zemel}]{PN}
Jake Snell, Kevin Swersky, and Richard Zemel. 2017.
\newblock Prototypical networks for few-shot learning.
\newblock In \emph{Proc. of NIPS}, volume~30, pages 4077--4087.

\bibitem[{Vinyals et~al.(2016)Vinyals, Blundell, Lillicrap, Wierstra
  et~al.}]{vinyals2016matching}
Oriol Vinyals, Charles Blundell, Timothy Lillicrap, Daan Wierstra, et~al. 2016.
\newblock Matching networks for one shot learning.
\newblock In \emph{Proc. of NIPS}, volume~29, pages 3630--3638.

\bibitem[{Wolf et~al.(2019)Wolf, Debut, Sanh, Chaumond, Delangue, Moi, Cistac,
  Rault, Louf, Funtowicz et~al.}]{wolf2019huggingface}
Thomas Wolf, Lysandre Debut, Victor Sanh, Julien Chaumond, Clement Delangue,
  Anthony Moi, Pierric Cistac, Tim Rault, R{\'e}mi Louf, Morgan Funtowicz,
  et~al. 2019.
\newblock Huggingface's transformers: State-of-the-art natural language
  processing.
\newblock \emph{arXiv preprint arXiv:1910.03771}.

\bibitem[{Xu and Xiang(2021)}]{Frog-GNN}
Shiyao Xu and Yang Xiang. 2021.
\newblock Frog-gnn: Multi-perspective aggregation based graph neural network
  for few-shot text classification.
\newblock \emph{Expert Systems with Applications}, 176:114795.

\bibitem[{Yao et~al.(2021)Yao, Wu, Al-Shedivat, and Xing}]{KGML}
Huaxiu Yao, Ying-xin Wu, Maruan Al-Shedivat, and Eric Xing. 2021.
\newblock Knowledge-aware meta-learning for low-resource text classification.
\newblock In \emph{Proc. of {EMNLP}}, pages 1814--1821.

\bibitem[{Ye and Ling(2019)}]{MLMAN}
Zhi-Xiu Ye and Zhen-Hua Ling. 2019.
\newblock Multi-level matching and aggregation network for few-shot relation
  classification.
\newblock In \emph{Proc. of {ACL}}, pages 2872--2881.

\bibitem[{Zhao et~al.(2021)Zhao, Wallace, Feng, Klein, and
  Singh}]{zhao2021calibrate}
Zihao Zhao, Eric Wallace, Shi Feng, Dan Klein, and Sameer Singh. 2021.
\newblock Calibrate before use: Improving few-shot performance of language
  models.
\newblock In \emph{International Conference on Machine Learning}, pages
  12697--12706.

\bibitem[{Zhong et~al.(2021{\natexlab{a}})Zhong, Lee, Zhang, and
  Klein}]{zhong2021adapting}
Ruiqi Zhong, Kristy Lee, Zheng Zhang, and Dan Klein. 2021{\natexlab{a}}.
\newblock Adapting language models for zero-shot learning by meta-tuning on
  dataset and prompt collections.
\newblock In \emph{Findings of EMNLP}, pages 2856--2878.

\bibitem[{Zhong et~al.(2021{\natexlab{b}})Zhong, Friedman, and
  Chen}]{zhong2021fact}
Zexuan Zhong, Dan Friedman, and Danqi Chen. 2021{\natexlab{b}}.
\newblock Factual probing is {[MASK]:} learning vs. learning to recall.
\newblock In \emph{Proc. of {NAACL-HLT}}, pages 5017--5033.

\end{thebibliography}
\bibliographystyle{acl_natbib}

\appendix
\section{Full experiment results}
\label{appendix: A}
Here we provide full experiment results with standard deviation. Full results are shown in Table~\ref{tbl:full}.

\begin{sidewaystable*}[t]
\small
\begin{tabular}{l cc cc cc cc cc}
\toprule
\multicolumn{1}{c}{Method} &
\multicolumn{2}{c}{20 News} &
\multicolumn{2}{c}{Amazon} &
\multicolumn{2}{c}{HuffPost} &
\multicolumn{2}{c}{Reuters} &
\multicolumn{2}{c}{Average}
\\
\cmidrule(lr{0.5em}){2-3}\cmidrule(lr{0.5em}){4-5}\cmidrule(lr{0.5em}){6-7}\cmidrule(lr{0.5em}){8-9}\cmidrule(lr{0.5em}){10-11}
 & 1 shot & 5 shot & 1 shot & 5 shot & 1 shot & 5 shot & 1 shot & 5 shot & 1 shot & 5 shot 
\\
\midrule
\textsc{1-NN}       & $38.8$ &  $51.9$ & $51.4$ &  $67.1$ & $31.5$ &  $42.3$ & $ 57.8 $ &  $ 82.9 $ & $ 44.88 $ &  $ 61.05 $\\
\textsc{FT}        & $33.0$ &  $47.1$ & $45.7$ &  $63.9$ & $32.4$ &  $44.1$ & $ 70.9 $ &  $ 91.0 $ & $ 45.50 $ &  $ 61.53 $\\
\textsc{Proto}     & $37.8$ &  $46.5$ & $41.9$ &  $59.2$ & $34.8$ &  $50.2$ & $ 61.0 $ &  $ 72.1 $  & $43.88$ &  $57.00$\\
\textsc{MAML}      & $37.2$ & $48.6$ & $43.6$ & $62.4$ & $38.9$ & $53.7$ & $ 61.5 $ & $72.0$ & $45.30$ & $59.18$ \\
\textsc{RR}        & $44.8$ &  $64.3$ & $60.2$ &  $79.7$ & $37.6$ &  $59.5$ & $69.1$ &  $93.0$ & $52.93$ &  $74.13$\\
\textsc{DS}~\citeyearpar{Few-shotTC} &  $52.1$ & $68.3$ & $62.6$ & $81.1$ & $43.0$ & $63.5$ & $81.8$ & $96.0$ & $59.88$ & $77.23$\\
\textsc{DE-MLMAN}~\citeyearpar{DE}        & $-$ &  $-$ & $-$ & $- $ & $49.7$ &  $60.9$ & $-$ &  $-$ & $-$ &  $-$\\
\textsc{DE-MAML}~\citeyearpar{DE}         & $-$ &  $-$ & $-$ & $- $ & $51.8$ &  $67.3$ & $-$ &  $-$ & $-$ &  $-$\\
\textsc{DE-Proto}~\citeyearpar{DE}        & $-$ &  $-$ & $-$ & $- $ & $52.3$ &  $69.6$ & $-$ &  $-$ & $-$ &  $-$\\
\textsc{KGML-Proto}~\citeyearpar{KGML}        & $-$ &  $-$ & $58.6$ &  $74.5$ & $42.3$ &  $58.7$ & $-$ &  $-$ & $-$ &  $-$\\
\textsc{KGML-MAML}~\citeyearpar{KGML}        & $-$ &  $-$ & $51.4$ &  $58.8$ & $44.2$ &  $54.1$ & $-$ &  $-$ & $-$ &  $-$\\
\textsc{P-Tuning}~\citeyearpar{P-tuning}        & $61.20$ &  $71.47$ & $62.18$ &  $79.13$ & $54.48$ &  $65.75$ & $90.01$ &  $96.66$ & $66.97$ &  $78.25$\\
\textsc{Frog-GNN}~\citeyearpar{Frog-GNN}        & $-$ &  $-$ & $71.5$ &  $83.6$ & $54.1$ &  $69.6$ & $-$ &  $-$ & $-$ &  $-$\\
\textsc{LaSAML-PN}~\citeyearpar{LaSAML}         & $-$ &  $-$ & $-$ & $- $ & $62.1$ &  $70.1$ & $-$ &  $-$ & $-$ &  $-$\\
\textsc{ContrastNet}~\citeyearpar{chen2022contrastnet}         & $71.74$ &  $\bm{81.57}$ & $76.13$ & $85.17$ & $53.06$ &  $65.32$ & $86.42$ &  $95.33$ & $71.84$ &  $81.85$\\

\midrule
\textsc{Ours~(Pretrain Init)} & $72.52$\tiny{$\pm{0.67}$}  & $76.32$\tiny{$\pm{0.21}$} & $75.12$\tiny{$\pm{0.65}$} & $83.27$\tiny{$\pm{0.03}$} & $70.82$\tiny{$\pm{0.74}$} & $75.47$\tiny{$\pm{0.24}$} & $95.07$\tiny{$\pm{0.15}$} & $\bm{97.29}$\tiny{$\pm{0.06}$} & $78.38$ & $83.09$\\
\textsc{Ours~(Meta Init)} & $\bm{73.75}$\tiny{$\pm{0.10}$}  & $76.57$\tiny{$\pm{0.06}$} & $\bm{77.65}$\tiny{$\pm{0.47}$} & $\bm{85.54}$\tiny{$\pm{0.03}$} & $\bm{71.93}$\tiny{$\pm{0.22}$} & $\bm{76.32}$\tiny{$\pm{0.05}$} & $\bm{95.20}$\tiny{$\pm{0.12}$} & $97.17$\tiny{$\pm{0.18}$} & $\bm{79.63}$ & $\bm{83.90}$\\
\bottomrule
\end{tabular}
\centering
\caption{Full results of 1-shot and 5-shot classification on four datasets in terms of accuracy. The rows below the mid-line are results of MetaPrompting. ‘-’ means that the result of this dataset is not given in the original paper.}
\label{tbl:full}
\vspace*{-3mm}
\end{sidewaystable*}

\end{document}